# A Dynamic Approach to Probabilistic Inference using Bayesian Networks

**Michael C. Horsch** and **David Poole***
Department of Computer Science
University of British Columbia
Vancouver, British Columbia
Canada
email: horsch@cs.ubc.ca, poole@cs.ubc.ca

## Abstract

In this paper we present a framework for dynamically constructing Bayesian networks. We introduce the notion of a background knowledge base of *schemata*, which is a collection of parameterized conditional probability statements. These schemata explicitly separate the general knowledge of properties an individual may have from the specific knowledge of particular individuals that may have these properties. Knowledge of individuals can be combined with this background knowledge to create Bayesian networks, which can then be used in any propagation scheme.

We discuss the theory and assumptions necessary for the implementation of dynamic Bayesian networks, and indicate where our approach may be useful.

## 1 Motivation

Bayesian networks are used in AI applications to model uncertainty and perform inference. They are often used in expert systems [Andreassen *et al.*, 1987], and decision analysis[Schachter, 1988, Howard and Matheson, 1981], in which the network is engineered to perform a highly specialized analysis task.

A Bayesian network often implicitly combines general knowledge with specific knowledge. For example, a Bayesian network with an arc as in Figure 1 refers to a specific individual (a house or a tree or dinner or whatever), exhibiting a somewhat generalized property (fire causes smoke).

Our dynamic approach is motivated by the observation that a knowledge engineer has expertise in a domain, but may not be able to anticipate the individuals in the model. By separating properties from individuals the knowledge engineer can write a knowledge base which is independent of the individuals; the system user can tell the system which individuals to consider, because she can make this observation at run time. The system user doesn't have to be an expert in the domain, to create an appropriate network.

As an example, suppose we are using Bayesian networks to model probabilistically the response of several people to the sound of an alarm. Our approach allows the observation of any number of people. The same knowledge about how people respond to alarms is used for each person.

There are two parts to our approach. First, we provide a collection of schemata, which are parameterized, and can be used when necessary given the details of the problem. In particular, the same piece of knowledge may be instantiated several times in a single dynamically created network.

Second, an automatic process builds a Bayesian network by combining the observation of individuals with the schemata. Thus, if we want to reason about a situation involving a fire alarm, given that three different people all hear the same alarm, this information, provided as evidence to our inference engine, causes the appropriate network to be created.

The Bayesian network constructed dynamically can absorb evidence (conditioning) to provide posterior probabilities using any propagation scheme[Pearl, 1988, Lauritzen and Spiegelhalter, 1988, Schachter, 1988].

We now proceed with a cursory introduction to Bayesian networks, followed by a presentation of our dynamic approach, giving some examples of how dynamic networks can be used. Finally, we draw some conclusions concerning the applicability of dynamic networks to particular domains.

## 2 Bayesian Networks

A Bayesian network is a directed acyclic graph which represents in graphical form the joint probability distribution, and the statistical independence assumptions, for a set of random variables.

A node in the graph represents a variable, and an arc indicates that the node at the head of the arc is directly dependent on, or *conditioned by* the node at

*This research is supported in part by NSERC grant #OGPOO44121.

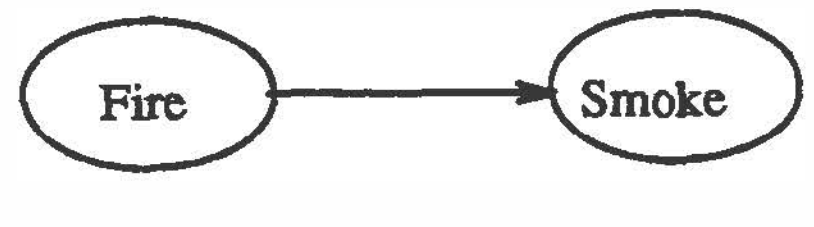

Figure 1: A simple arc implicitly representing an individual.

the tail. The collection of arcs directed to a variable give the independence assumptions for the dependent variable. Associated with this collection is a prior probability distribution, or ***contingency table***, which quantifies the effects of observing events for the conditioning variables on the probability of the dependent variable.

The graphical representation is used in various ways to calculate posterior joint probabilities. Pearl [Pearl, 1988] uses the arcs to propagate causal and diagnostic support values throughout a singly-connected network. Lauritzen and Spiegelhalter [Lauritzen and Spiegelhalter, 1988] perform evidence absorption and propagation by constructing a triangulated graph based on the Bayesian network that models the domain knowledge. Schachter [Schachter, 1988] uses the arcs to perform node reduction on the network.[1] Poole and Neufeld [Poole and Neufeld, 1989] implement an axiomatization of probability theory in Prolog which uses the arcs of the network to calculate probabilities using "reasoning by cases" with the conditioning variables. Our current implementation uses the work of Poole and Neufeld, but is not dependent on it.

## 3 Representation Issues

Creating networks automatically raises several issues which we discuss in this section. First, we need to represent our background knowledge in a way which preserves a coherent joint distribution defined by Bayesian networks, and which also facilitates its use in arbitrary situations. Furthermore, there are situations in which using the background knowledge may lead to ambiguity, and our approach must also deal with this.

Before we get into our discussion, a short section dealing with syntactic conventions will help make things clearer.

### 3.1 Some syntax

In this section we clarify the distinction between our background knowledge and Bayesian networks.

[1]Schachter's influence diagrams contain decision and test nodes, as well as other devices not being considered in this paper. We are only looking at probabilistic nodes at present.

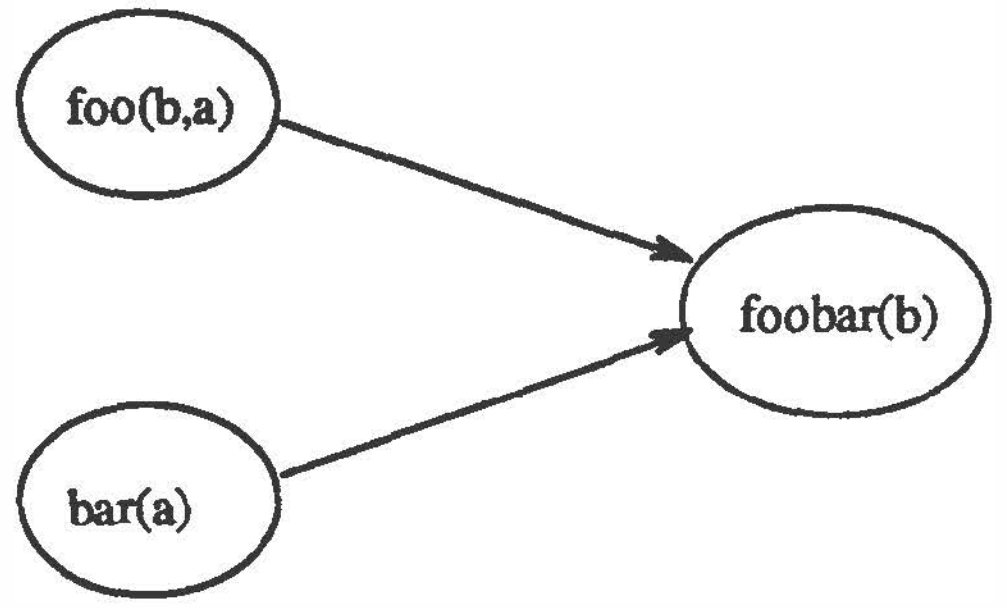

Figure 2: A simple instantiation of a schema.

In our Bayesian networks, random variables are propositions written like ground Prolog terms. For example, the random variable *flies(e127)* could represent the proposition that individual *e127* flies.

A ***schema*** is part of the background knowledge base, and describes qualitatively the direct dependencies of a random variable. In this paper, a schema is stated declaratively in sans serif font.

A schema can be defined constructively: a parameterized atom is written as a Prolog term with parameters capitalized. A schema is represented by the following:

$$a_1, \ldots, a_n \longrightarrow b$$

where the arrow $\longrightarrow$ indicates that the parameterized atom, $b$, on the left side is directly dependent on the parameterized atoms, $a_i$, on the right.

An instantiation of a parameter is the substitution of a parameter by a constant representing an individual. We indicate this in our examples by listing the individuals in a set, as in {e127}. Instantiating a parameterized atom creates a proposition that can be used in a Bayesian network.

A schema is instantiated when all parameters have been instantiated, and an instantiated schema becomes a part of the Bayesian network, indicating a directed arc from the instantiation of each $a_i$ and the instantiation of $b$.

For example, the schema:

foo(X,a), bar(a) ⟶foobar(X)

with {b} instantiating the parameter X, creates the network[2] shown in Figure 2.

The schemata in the background knowledge base are isolated pieces of information, whereas a Bayesian network is a single entity. An interpretation is that a Bayesian network defines a joint distribution, whereas the schemata in a knowledge base define conditional probability factors of which instances are used to compute a joint probability distribution.

[2]In our examples, Bayesian networks are pictured as encircled node labels connected by directed arrows, as in Figure 2.

A schema is incomplete without the contingency table which quantifies the connection between the dependent variable and its conditioning parents.[3] These tables have the following form, and follow the same syntactic assumptions as the symbolic schemata:

| | | |
|---|---|---|
| p(foobar(X) \| foo(X,a), bar(a)) | = | 0.95 |
| p(foobar(X) \| foo(X,a), ¬bar(a)) | = | 0.666 |
| p(foobar(X) \| ¬foo(X,a), bar(a)) | = | 0.25 |
| p(foobar(X) \| ¬foo(X,a), ¬bar(a)) | = | 0.15 |

By writing a parameterized schema, we are licensing a construction process to instantiate the parameters in the schema for any individual. This requires that the knowledge we provide as schemata must be written generally.

The parameters in the schema and the contingent probability information should not be interpreted with any kind of implicit quantification. Rather, they are intended to apply to a "typical" individual. Furthermore, when this parameterized information is used, the parameters should be replaced with constants which do not convey any intrinsic information. By designing schemata carefully, we can avoid the "Ace of Spades" problem [Schubert, 1988].

## 4 Dynamic Instantiation

A general schema with parameters can be instantiated for more than one individual. There are three cases:

### 4.1 Unique schemata

These are schema which have no parameters, or which have the same parameters occurring on both sides of the arc. For example:

$$a(X), b \longrightarrow c(X).$$

For the constants {x,y} instantiating the parameter X, the network building process puts two arcs in our network, as in Figure 3.

### 4.2 Right–multiple schemata

In the second case, there are parameters on the right side of the schema which do not occur on the left side, as in a,b ⟶c(Y). (Note that, in addition, there may be parameters which occur on both sides of the arc). Instantiating the parameter Y in the above schema with values {x, y, z} creates the network shown in Figure 4.

**Example 4–1:** Pearl [Pearl, 1988, Chapter 2] presents an extended example, about burglar alarms and testimonies from several people, as motivation for the use of Bayesian networks in an automated reasoning system. The following schemata can specify this problem:

| | | |
|---|---|---|
| burglary, earthquake | ⟶ | alarm_sound |
| earthquake | ⟶ | news_report |
| alarm_sound | ⟶ | testimony(X) |
| alarm_sound | ⟶ | call(Y) |

Note the use of right–multiple schemata. This allows us to specify multiple cases for testimonies from different people. In particular, we could tell the system that we are investigating the hypothesis that a burglary occurred based on testimonies from Dr Watson and Mrs Gibbons. people, by supplying the constants representing these individuals at run time. □

[3] In our examples, we do not provide the table explicitly, we assume its existence implicitly.

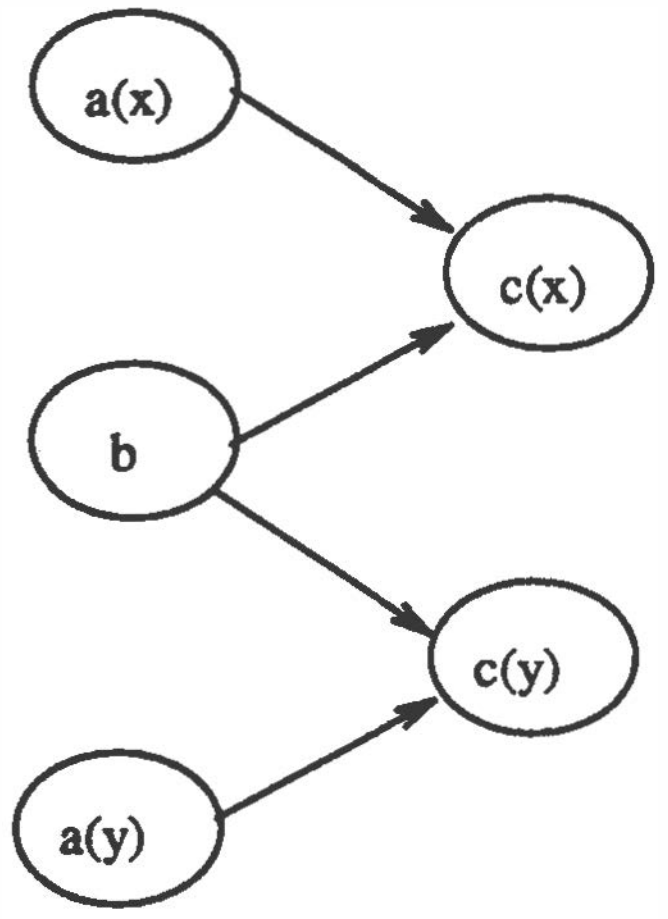

Figure 3: The network created by instantiating the parameterized unique schemata from Section 4.1.

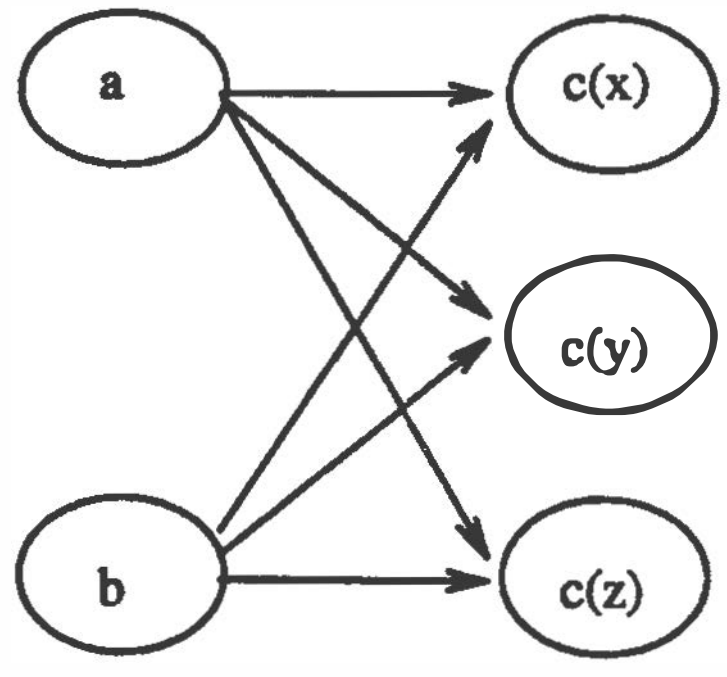

Figure 4: The network created by instantiating the parameterized right–multiple schema from Section 4.2.

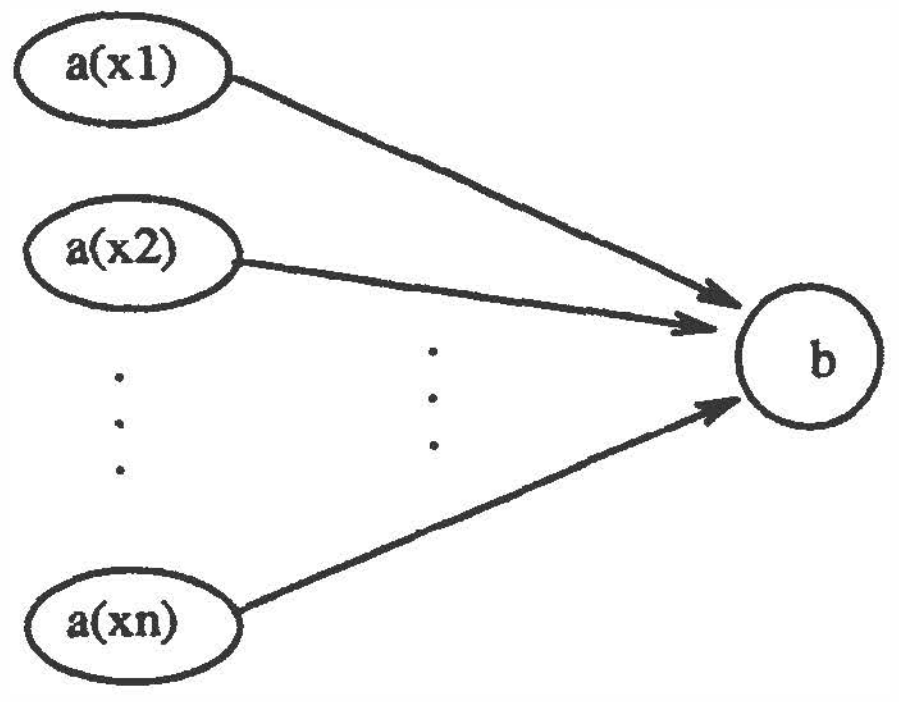

Figure 5: The network created by instantiating the parameterized left–multiple schema from Section 4.3.

### 4.3 Left–multiple schemata

The remaining case is characterized by parameters occurring on the left side of the schema which do not occur on the right side. For example:

$$a(X) \longrightarrow b$$

Instantiating the parameter X with a number of individuals {x1, ..., xn} creates a network structure as shown in Figure 5. The conditional probability for b must be stated in terms of multiple instances of the node a(X). To use the multiple instantiation of a left–multiple schema, we need to be able to provide contingencies for any arbitrary number of individuals. This is problematic because it is not reasonable to provide this kind of information (in terms of contingency tables, for example), as background knowledge: we would require a table for every possible set of individuals. To be useful, the contingency tables for left–multiple schemata must allow for any number of instantiations.

It would be representationally restrictive to disallow the use of left–multiple schemata, because sometimes we do want to allow an unspecified number of variables to condition a dependent variable. For example, we may want to include in our knowledge base the idea that any number of people may call the fire department to report a fire.

We provide two mechanisms, called existential and universal combination, which specify how an unknown number of possible conditioning variables affects another variable. These are based on the *Canonical Models of Multicausal Interactions* suggested by Pearl [Pearl, 1988].

#### Existential Combination

Consider the following example: A person who smells smoke may set off a fire alarm, and the sound of the fire alarm may cause others to leave the building.

This situation shows the usefulness of an existential combination over individuals in the problem. We want to combine the effects of each individual into a single variable which relates the likelihood that at least one of the individuals satisfied the condition. The following schema represents this notion syntactically:

$$\exists X \in type \cdot a(X) \longrightarrow b\,.$$

The existential schema serves as a short–hand notation for a dynamic Or–rule structure. For example, when it is given that {x1, x2, ..., xn} are members of the set type, the schema above expands into a network structure as in Figure 6, where the contingency table for ∃X∈type·a(X) is unity if any of a(X) is known to be true, and zero if they are all false. The Or–rule is indicated in the diagram by the circular arc on the inputs to the existential node.

This schema requires a contingency table such as:

$$p(b \mid \exists X \in type \cdot a(X)) = 0.7665$$
$$p(b \mid \neg \exists X \in type \cdot a(X)) = 0.0332$$

which describes the effect on b for the case where at least one known individual satisfies a(X), and the case where no known individual does (these numbers are arbitrary).

**Example 4–2:** The example about fire alarms, given at the beginning of this section, can be represented by the following background knowledge base:

| | | |
|---|---|---|
| fire | ⟶ | smells_smoke(X) |
| smells_smoke(X) | ⟶ | sets_off_alarm(X) |
| ∃Y∈person·sets_off_alarm(Y) | ⟶ | alarm_sounds |
| alarm_sounds | ⟶ | leaves_building(Z) |

Suppose we are given that *john* and *mary* are the only known members of the set *person*. This information creates the network shown in Figure 7. □

There are several points which should be made:

1. The variable ∃X∈type·a(X) is propositional, and when it is true, it should be interpreted as the fact that some known individual satisfies the proposition a(X). When this existential combination is false, the interpretation should be that no known individual satisfies the proposition a(X), although there may be an individual satisfying the proposition who is unknown to the system.
2. The proposition ∃X∈type·a(X) can be considered a schema on it's own, acting like a variable which is disjunctively influenced by a(c) for every constant c in the set type. The intermediate Bayesian arcs are not written as part of the knowledge base.
3. The set type, from which the individuals for this schema are taken, serves two purposes: to constrain the applicability of the individuals to the schema, and to facilitate the dynamic specification of individuals. In this way, all and only those individuals who are known to satisfy the type requirement will be included in this combination.

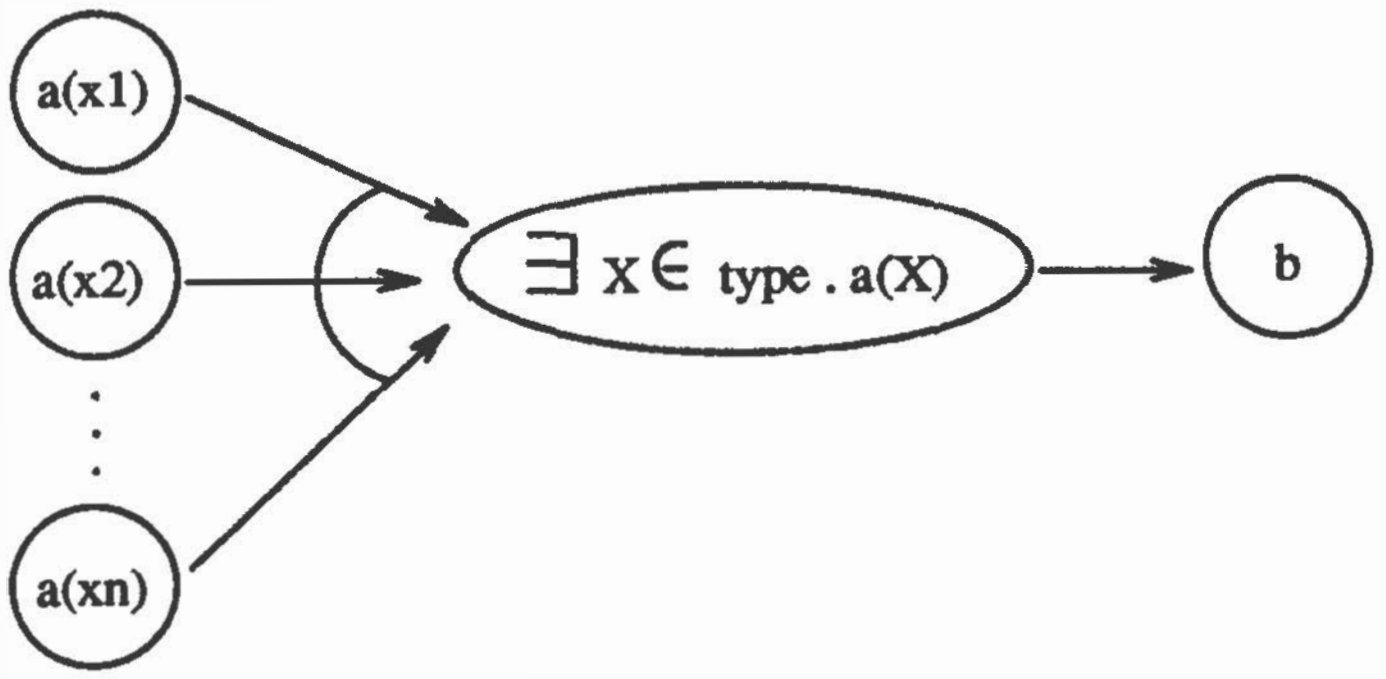

Figure 6: The instantiation of the existential combination node from Section 4.3.

**Universal Combination**

Universal combination can be used in situations for which all individuals of a certain type must satisfy a condition to affect a consequent. For example, a board of directors meeting can begin only when all the members who are going to come are present.

We represent this idea with a schema with the following syntax:

$$\forall X \in \mathsf{type} \cdot \mathsf{a}(X) \longrightarrow \mathsf{b}$$

The discussion concerning the existential combination applies here, with the only exception being the interpretation of the schema. The variable $\forall X \in \mathsf{type} \cdot \mathsf{a}(X)$ is true if $\mathsf{a}(X)$ is true for every member of the set type. We treat the combination as a dynamic And–rule over all members of the set type. The And–rule contingency table is unity if all the members satisfy the condition $\mathsf{a}(X)$, and zero otherwise.

**Example 4–3:** A board meeting for a large corporation requires the presence of all board members, and the meeting may result in some actions, say, buying out a smaller company. A board member may attend the meeting, depending on her reliability and state of health.

$$\begin{array}{lcl} \forall X \in \mathsf{board_members} \cdot \mathsf{present}(X) & \longrightarrow & \mathsf{meeting} \\ \mathsf{meeting} & \longrightarrow & \mathsf{buy_out} \\ \mathsf{sick}(X), \mathsf{reliable}(X) & \longrightarrow & \mathsf{present}(X) \end{array}$$

We note that at the time the schemata are written, it is not important to know how many board members there may be. However, we must know exactly who is a member on the board before the network can be created. This information is supplied at run time, and the construction process can construct the appropriate network. Evidence concerning the reliability and health of any board member can then be submitted to the network, providing appropriate conditioning for queries. □

## 5 Creating Bayesian networks

To create a Bayesian network, the schemata in our knowledge base must be combined with the individuals the process knows about. The process by which this is accomplished in our implementation has been kept simple, and syntactic.

Every schema in the knowledge base is instantiated by substituting ground atoms representing the individuals known to be part of the model. Each instantiation of a schema is an arc in the Bayesian network. The contingency table associated with each schema is also instantiated, providing the prior probability information we require.

This procedure creates the collection of Bayesian arcs we use as the Bayesian network for our model.

## 6 Current and Future Work

These proceedings report simultaneous and independent research on building Bayesian networks dynamically. Goldman and Charniak [Goldman and Charniak, 1990] use a sophisticated rule base to construct Bayesian networks using more 'semantic' knowledge. As well, their approach handles cases where arcs are added to a network which has already absorbed evidence.

We have outlined only two ways in which the effects of an unknown number of individuals can be combined. These could be extended to include other types of combination.

This paper describes how knowledge of individuals can be used to create a Bayesian network. There are times when we want to use observations about the individuals to add arcs to the network. For instance, observing the topology of an electronic circuit should help structure the Bayesian network which models the uncertainty in diagnosing faults. These observations have a unusual property: they can be used to structure a network if they are observed to be true, and yield no exploitable structural information if they are observed false. We are looking at how we might implement this idea.

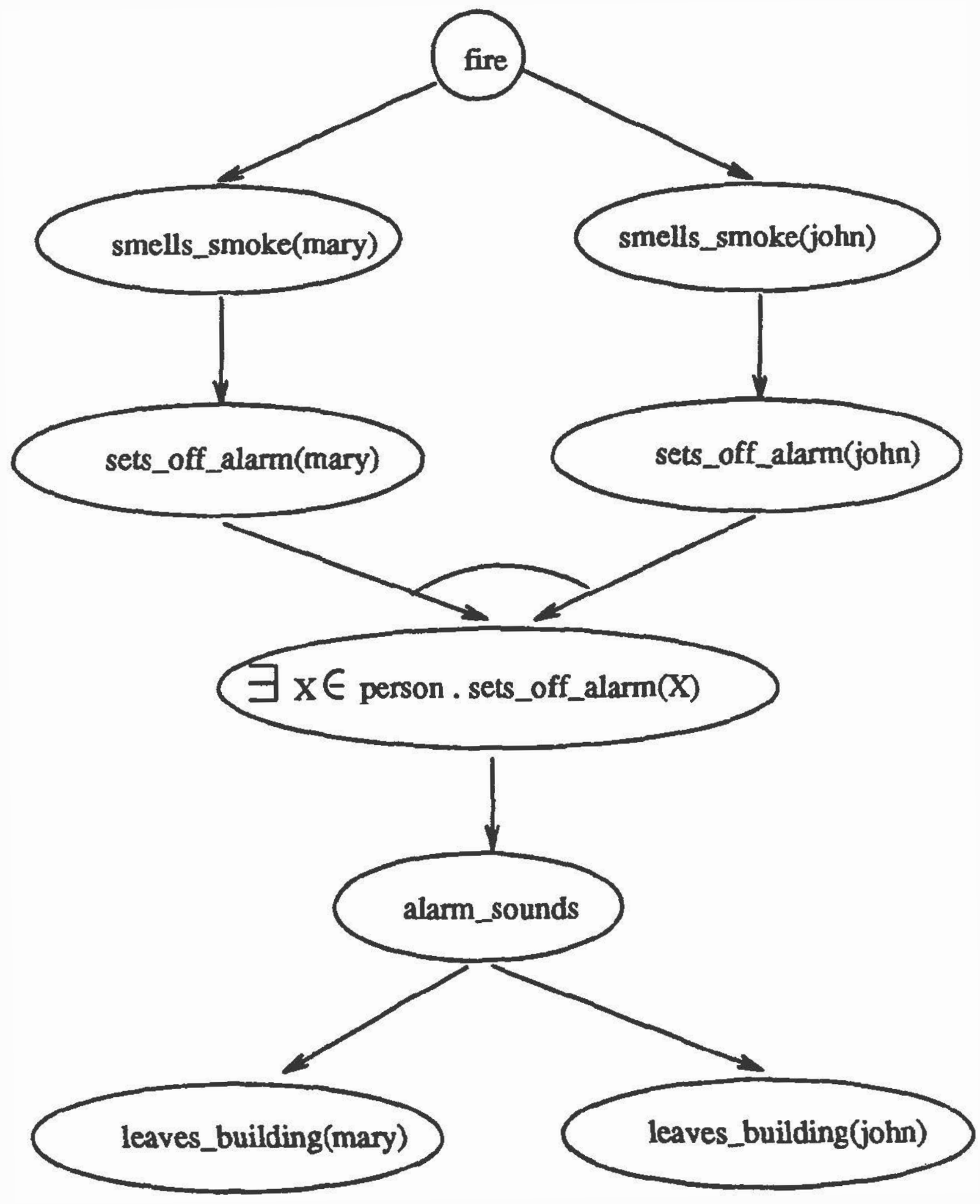

Figure 7: The Bayesian network created for Example 4–2

The process, described in this paper, which creates the Bayesian network from the knowledge base is very simple, and many improvements could be made. One way to limit the size of our networks is to delay the creation of the network until all the evidence available has been observed. This would allow the exploitation of conditional independence at the level of the structure of the network, making the network smaller and easier to evaluate. This idea is not exploited by many systems (for example, Lauritzen and Spiegelhalter [Lauritzen and Spiegelhalter, 1988] mention the idea, but argue against it).

The ideas in this paper have been implemented and tested using toy problems. Currently, larger applications are being built in more sophisticated domains, including circuit diagnosis and the interpretation of sketch maps.

## 7 Conclusions

In this paper we have presented a dynamic approach to the use of Bayesian networks which separates background and specific knowledge. Bayesian networks are created by combining parameterized schemata with the knowledge of individuals in the model.

Our approach to the use of Bayesian networks has been from the viewpoint of providing an automated tool for probabilistic reasoning, and is useful for modelling problems for domains in which many details of the problem cannot be exhaustively anticipated by the designer of the probabilistic model. Such domains include expert systems, diagnosis, story understanding, natural language processing and others.

An implementation of this approach has been completed in Prolog.